\documentclass[runningheads]{llncs}
\usepackage[T1]{fontenc}
\usepackage{graphicx}
\usepackage{booktabs}
\usepackage{algorithm}
\usepackage{algorithmic}
\usepackage[switch]{lineno}
\usepackage{textcomp}
\usepackage{amsmath,amssymb,amsfonts}
\usepackage{enumitem}
\usepackage{xcolor}
\usepackage{color}
\usepackage{colortbl}
\usepackage{times}
\usepackage{soul}
\usepackage{url}
\usepackage[hidelinks]{hyperref}
\usepackage[utf8]{inputenc}
\usepackage[small]{caption}
\usepackage{graphicx}
\usepackage{multirow}
\usepackage{amsmath}
\usepackage[misc]{ifsym}

\usepackage{mwe}

\begin{document}

\title{Frequency Enhanced Pre-training for Cross-city Few-shot Traffic Forecasting}
\toctitle{Frequency Enhanced Pre-training for Cross-city Few-shot Traffic Forecasting}

\author{Zhanyu~Liu \and
Jianrong~Ding \and
Guanjie~Zheng(\Letter\footnotetext{\Letter: Corresponding Author})
}
\tocauthor{Zhanyu~Liu, Jianrong~Ding, Guanjie~Zheng}
\authorrunning{Z. Liu et al.}
\institute{
Shanghai Jiao Tong University, Shanghai, China
\email{\{zhyliu00,rafaelding,gjzheng\}@sjtu.edu.cn
}
}

\maketitle              



\definecolor{lgray}{rgb}{0.9,0.9,0.9}

\begin{abstract}
The field of Intelligent Transportation Systems (ITS) relies on accurate traffic forecasting to enable various downstream applications.
However, developing cities often face challenges in collecting sufficient training traffic data due to limited resources and outdated infrastructure.
Recognizing this obstacle, the concept of cross-city few-shot forecasting has emerged as a viable approach.
While previous cross-city few-shot forecasting methods ignore the frequency similarity between cities,
we have made an observation that the traffic data is more similar in the frequency domain between cities.
Based on this fact, we propose a \textbf{F}requency \textbf{E}nhanced \textbf{P}re-training Framework for \textbf{Cross}-city Few-shot Forecasting (\textbf{FEPCross}). 
FEPCross has a pre-training stage and a fine-tuning stage.
In the pre-training stage, we propose a novel Cross-Domain Spatial-Temporal Encoder that incorporates the information of the time and frequency domain and trains it with self-supervised tasks encompassing reconstruction and contrastive objectives.
In the fine-tuning stage, we design modules to enrich training samples and maintain a momentum-updated graph structure, thereby mitigating the risk of overfitting to the few-shot training data.
Empirical evaluations performed on real-world traffic datasets validate the exceptional efficacy of FEPCross, outperforming existing approaches of diverse categories and demonstrating characteristics that foster the progress of cross-city few-shot forecasting.
\end{abstract}
\section{Introduction}

\begin{figure*}
    \centering
    \includegraphics[width=0.94\linewidth]{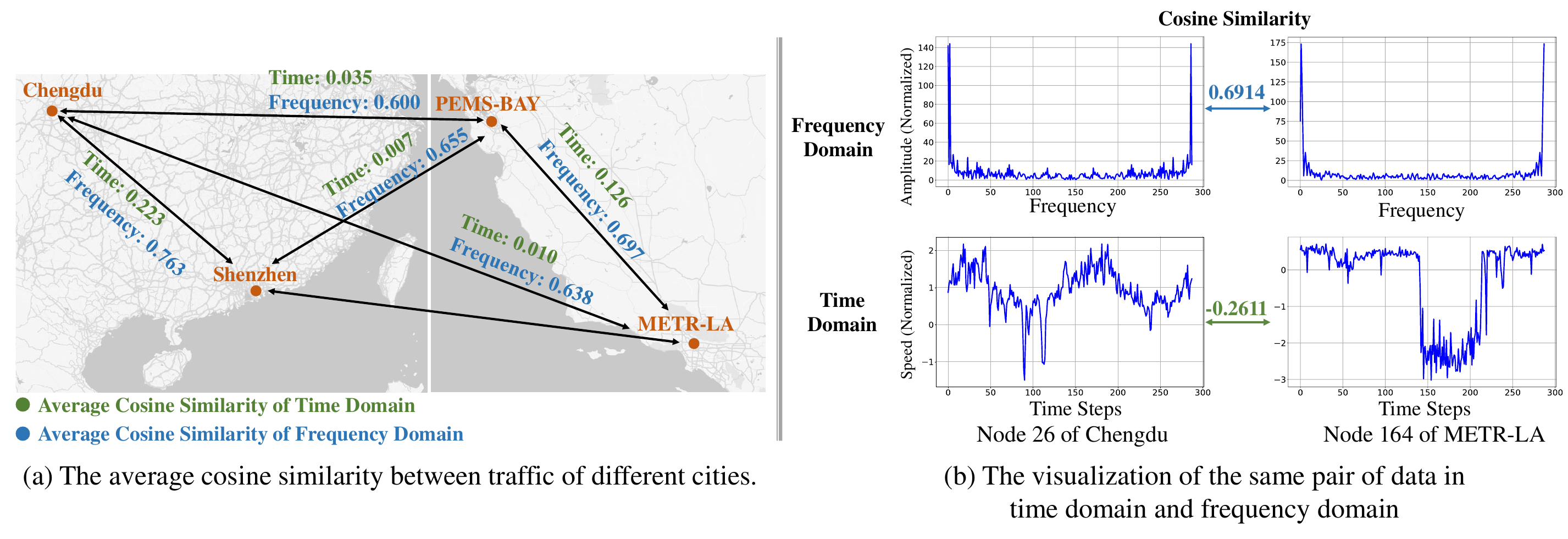}
    \caption{(a) The mean cosine similarity of one-week traffic speed data within the time and frequency domains across four cities. We can observe a higher level of similarity in the frequency domain.
    (b) An illustrative instance showcasing a pair of data in both the time domain and frequency domain reflects that, despite containing identical data, the frequency domain exhibits a significantly higher level of cosine similarity.}
    \label{fig:1intro}
\end{figure*}

The field of Intelligent Transportation Systems (ITS) has recognized the significance of traffic forecasting as a critical service.
By effectively employing historical traffic data to precisely forecast future traffic, it becomes feasible to enable diverse downstream applications, including traffic signal control~\cite{wei2018intellilight} and traffic tolling~\cite{wang2022ctrl}.
In order to gather the necessary traffic data for training a traffic forecasting model, the utilization of traffic sensors~\cite{li2017diffusion} or vehicle devices~\cite{Didi} is crucial.
The devices are well deployed in the developed cities and the data collection is easy.
However, when it comes to developing cities, one of the main obstacles is the lack of adequate infrastructure to support the deployment of these devices. 
Developing cities often have limited resources and outdated technology systems, making it difficult to establish a reliable network infrastructure for data collection.
Consequently, the traffic data in developing cities could be limited, and thus training a deep traffic forecasting model is hard.

To address the issue of limited data in developing cities and enhance the efficiency of traffic forecasting services, the concept of cross-city few-shot forecasting has emerged as a viable solution.
This approach involves learning from cities with abundant data and transferring that knowledge to cities with limited data.
In recent years, many works have focused on this problem.
RegionTrans~\cite{wang2018cross} and CrossTReS~\cite{jin2022selective} identify the inherent regional correlation between the source city and the target city.
MetaST~\cite{yao2019learning} learns a global memory which is then queried by the target region.
STrans-GAN~\cite{zhang2022strans} generates future traffic based on traffic demand with a GAN-based model.
However, these methods depend on auxiliary data, such as event information, for the transfer of knowledge. 
This reliance on auxiliary data poses a challenge when such data is not available in the city with limited data resources.
ST-MetaNet~\cite{pan2019urban} and ST-GFSL~\cite{lu2022spatio} utilize learned meta-knowledge to generate the parameters of the spatial-temporal neural network.
TransGTR~\cite{jin2023transferable} aims to learn the city-agnostic node features and the transferable graph structure between cities.
TPB~\cite{liu2023cross} aims to construct a traffic pattern bank that contains the meta-knowledge across cities.

Previous methods have predominantly concentrated on capturing temporal relationships between different cities.
However, they neglect the potential knowledge that can be derived and shared from the frequency domain.
In fact, as shown in Fig.~\ref{fig:1intro}, the traffic data between cities is more similar in the frequency domain.
We calculate the mean cosine similarity of aligned one-week traffic data of different cities in the time domain and the frequency domain, which is shown in Fig.~\ref{fig:1intro}(a).
We could observe that the traffic data exhibits notable similarity in the frequency domain across different cities.
Fig.~\ref{fig:1intro}(b) shows an example that even the same pair of traffic data exhibits substantial similarities in the frequency domain and they are both prominent in the low-frequency range.
The identification of these frequency domain patterns implies that modeling traffic data with a focus on the frequency domain can lead to a better understanding and capture of the underlying dynamics. 
By leveraging these patterns into the pre-training process in the data-rich source cities, it becomes possible to develop effective and accurate models for cross-city few-shot traffic forecasting.

Consequently, we propose a \textbf{F}requency \textbf{E}nhanced \textbf{P}re-training framework for \textbf{Cross}-city few-shot traffic forecasting, abbreviated as \textbf{FEPCross}. 
It contains two stages, the pre-training stage and the fine-tuning stage.
\textbf{For the pre-training stage}, we design a novel frequency-enhanced pre-training framework on the data-rich source city.
This framework incorporates the information from the time, amplitude, and phase domains through a Cross-Domain Spatial-Temporal Encoder.
By taking masked data from these three domains as input, the framework aims to reconstruct the missing information as a self-supervised task.
A contrastive loss is then added to guarantee the quality of the learned space.
\textbf{For the fine-tuning stage}, we design two modules based on the pre-trained Cross-Domain Spatial-Temporal Encoder to improve the performance of the few-shot fine-tuning process.
On one hand, we mask and reconstruct the few-shot training data to create more augmented training samples.
On the other hand, we maintain a momentum-updated graph structure of the target city.
These two modules not only improve the effectiveness but also enhance the resistance to the over-fitting of the fine-tuned model.

In summary, the main contributions are as follows.

\begin{itemize}
    \item We delve into the cross-city few-shot traffic forecasting task and find that the frequency domain pattern exhibits a higher degree of cross-city sharability. The potential of transferring knowledge from the frequency domain has been disregarded in prior research endeavors.
    \item We propose a Frequency Enhanced Pre-training framework (FEPCross) specifically designed for cross-city few-shot traffic forecasting. It incorporates
    information on various domains and contains modules to mitigate the risk of overfitting to the few-shot training data.
    \item We demonstrate the effectiveness of the FEPCross framework through extensive experiments on real-world traffic datasets from multiple cities.
    The results demonstrate that FEPCross achieves superior performances over the state-of-the-art baselines and exhibits distinctive characteristics such as prioritizing the modeling of low-frequency components, further contributing to its exceptional performance.
\end{itemize}

\section{Related Work}

\noindent
\textbf{Traffic Forecasting}
The practical applications of traffic forecasting have garnered significant attention.
Some work utilizes traditional statistical methods such as Kalman Filter~\cite{lippi2013short}, SVM~\cite{nikravesh2016mobile}, probalistic model~\cite{akagi2018fast}, or simulation~\cite{liang2022cblab}.
Some work combines modules such as GRU and GNN to model the spatial-temporal correlation within the traffic data, such as DCRNN~\cite{li2017diffusion}, STGCN~\cite{yu2017spatio}, 
GSTNet~\cite{fang2019gstnet}, LSGCN~\cite{huang2020lsgcn}, STFGNN~\cite{li2021spatial},
Frigate~\cite{gupta2023frigate},
HIEST~\cite{ma2023rethinking},
FDTI~\cite{liu2023fdti}.
To better capture the varying spatial-temporal correlations between traffic nodes, some work such as AGCRN~\cite{bai2020adaptive,duan2023localised}, Graph Wavenet~\cite{wu2019graph}, GMAN~\cite{zheng2020gman}, D2STGNN~\cite{shao2022decoupled}, ST-WA~\cite{cirstea2022towards}, DSTAGNN~\cite{lan2022dstagnn},
TrendGCN~\cite{jiang2023enhancing},
MC-STL~\cite{zhang2023mask} utilize adaptive methods such as node embedding to reconstruct the adaptive adjacent matrix and incorporate temporal long-term relations for improved predictions.
Some works utilize techniques to model the special characteristics of the traffic.
STG-NCDE~\cite{choi2022graph}, STDEN~\cite{ji2022stden} model the traffic based on ordinary differential equation (ODE).
FOGS~\cite{rao2022fogs} predict the first-order gradient of traffic flow.
STGBN~\cite{fan2023spatial} utilizes gradient boosting to enhance the model.
STEP~\cite{shao2022pre} adapts MAE and proposes a pipeline to pre-train a model.
Nevertheless, the aforementioned methods mainly focus on predicting traffic within a single city.
These methods would encounter difficulties such as distribution bias, the tendency to over-fit the single-city data, and the incompatibility of the node embedding technique in cross-city scenarios.

\begin{figure*}[!t]
    \centering
    \includegraphics[width=0.94\linewidth]{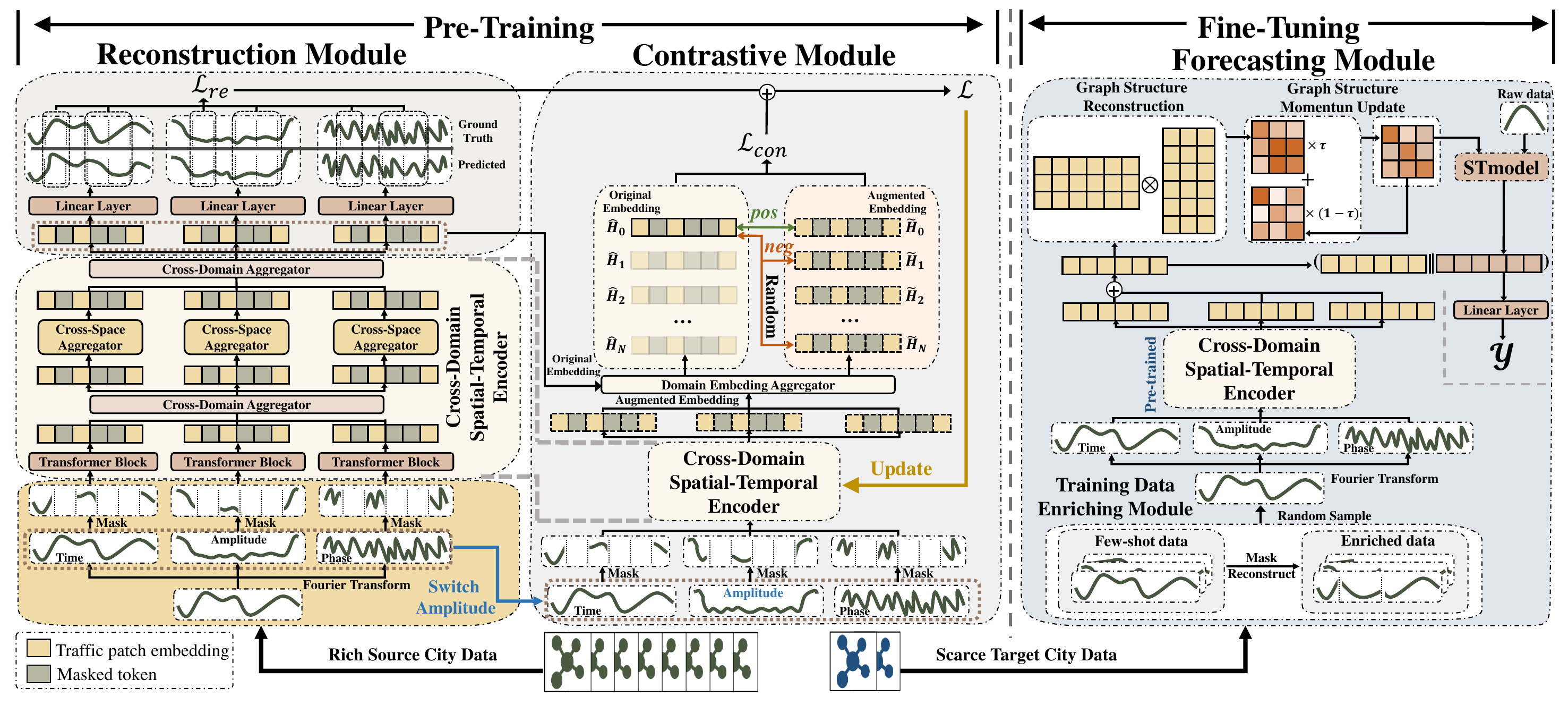}
    \caption{Overall Framework of FEPCross. }
    \label{fig:main_fig}
\end{figure*}

\noindent
\textbf{Cross-city Few-shot Forecasting}
To tackle the data scarcity problem and use the available knowledge of the data-rich cities effectively, several methods have been developed in the field of cross-city few-shot forecasting.
Floral~\cite{wei2016transfer} learns from the multimodal data from the source cities and transfers the knowledge to the target city for the air quality classification problem.
RegionTrans~\cite{wang2018cross}, CrossTReS~\cite{jin2022selective}, and MetaST~\cite{yao2019learning} learn the region correlation or the memory bank between the source cities to the target city.
However, these approaches utilize multimodal auxiliary information, which is not accessible in the data-scarce city.
ST-MetaNet~\cite{pan2019urban} and ST-GFSL~\cite{lu2022spatio} generate the parameters of spatial-temporal neural networks according to the learned meta-knowledge.
CityTrans~\cite{ouyang2023citytrans} uses domain adversarial training to transfer knowledge across cities.
UniST~\cite{yuan2024unist} uses prompt tuning to learn the spatial-temporal shared across cities.
TransGTR~\cite{jin2023transferable} learns city-agnostic features and transferable graphs from one city to another city.
TPB~\cite{liu2023cross} and MTPB~\cite{liu2024multi} construct the traffic pattern bank from multiple source cities and the data of the target city could query the bank to get the meta-knowledge to reconstruct the self-expressive graph structure~\cite{kang2022fine} and forecast the traffic.
However, these methods ignore the spectral similarity between the time series traffic data and thus get suboptimal results.

\section{Preliminary}

\textbf{Traffic Spatial-Temporal Graph:} 
The traffic spatial-temporal graph is the data structure that processes the traffic data.
It could be denoted as $\mathcal{G}=(\mathbf{V},\mathbf{A},\mathbf{X})$.
Here, $\mathbf{V}$ is the set of the nodes, and $N=|\mathbf{V}|$ is the number of nodes.
$\mathbf{A}\in\mathbb{R}^{N\times N}$ is the adjacency matrix that describes the correlation between nodes.
$\mathbf{X}\in\mathbb{R}^{N\times T\times C}$ is the traffic data of $N$ nodes, with time steps $T$ and channel $C$.

\noindent
\textbf{Traffic Forecasting:} The Traffic forecasting problem is to utilize the historical traffic to predict future traffic. Formally, given the historical traffic of $T_h$ steps, we aim to learn a model $f(\cdot)$ to forecast the future traffic of $T_f$ steps, which could be formulated as follows
\begin{equation}
    [\mathbf{X}_{t-T_h+1},\cdots,\mathbf{X}_{t}]\stackrel{f(\cdot)}{\longrightarrow}[\mathbf{X}_{t+1},\cdots,\mathbf{X}_{t+T_f}].
\end{equation}

\noindent
\textbf{Cross-city Few-Shot Traffic Forecasting:} 
Given a data-rich source city $\mathcal{G}^{source}$ and a data-scarce target city $\mathcal{G}^{target}$, the goal of cross-city few-shot traffic forecasting is to pre-train a model in $\mathcal{G}^{source}$ and fine-tune the model in $\mathcal{G}^{target}$. The fine-tuned model is expected to conduct accurate traffic forecasting on the future data of $\mathcal{G}^{target}$.
  

\section{Method}

In this section, we present the FEPCross framework, which comprises two stages: pre-training and fine-tuning as shown in Fig.~\ref{fig:main_fig}.
In the pre-training stage, we aim to learn robust traffic knowledge by the Cross-Domain Spatial-Temporal Encoder from the source city.
The encoder could capture the correlation across domains and spatial neighbors, and then reconstruct the masked input of three domains.
To encourage the encoder to learn a representation space that is both compact and discriminative, we incorporate a contrastive loss during the pre-training process.
In the fine-tuning stage, we address the challenges associated with limited data by enriching the few-shot training dataset of the target city using the pre-trained encoder.
This is achieved by applying a masking and reconstruction technique.
The graph structure is momentum-updated by the knowledge of the pre-trained encoder.
These two novel modules are designed to avoid over-fitting and improve the performance of the fine-tuned model.
Finally, a Spatial-Temporal model (STmodel) aggregates the short-term traffic knowledge and makes predictions.

\subsection{Pre-Training}
In this section, we will introduce the novel Cross-Domain Spatial-Temporal Encoder that captures the pattern of the traffic data across domains and spatial neighbors.
To train the Cross-Domain Spatial-Temporal Encoder, we employ two essential self-supervision tasks: reconstruction loss and contrastive loss. The reconstruction loss aims to reconstruct the original masked input data from the encoded representations. This task encourages the encoder to learn meaningful and informative representations that accurately capture the input data structure.

\noindent\textbf{Input:}
Fig.~\ref{fig:1intro} in the introduction section visually demonstrates that the frequency domain exhibits a higher similarity across different cities. 
Leveraging this observation, we adopt a Fourier Transform on the input traffic time series to extract the amplitude and phase domains. 
This transformation allows us to capture the frequency domain characteristics of the data.
Then, we segment the series into patches and randomly mask most of them.
Formally, by taking the time domain $Ti$ as the example, the input time series of node $i$, i.e., $\mathbf{X^{Ti}_{i, 0:T_h}}$ is separate to $P$ patches as $\mathbf{X^{Ti}_{i, 0:T_h}}=\{\mathbf{S^{Ti}_{i,0}}, \mathbf{S^{Ti}_{i,1}}, \cdots, \mathbf{S^{Ti}_{i,P}}\}$ and there is a masked vector $\mathbf{M}$ where $\mathbf{M^{Ti}_{i,k}=1}$ indicates $\mathbf{S^{Ti}_{i,k}}$ is masked.
The masked patch series of three domains is the input of the encoder similar to previous research~\cite{nie2022time,he2022masked}.

\noindent\textbf{Cross-Domain Spatial-Temporal Encoder:}
The Cros-Domain Spatial-Temporal Encoder aims to aggregate the information of the traffic data across domains and across neighbors.
Firstly, The unmasked part of the input is fed into a Transformer block.
Then, a learnable masked token $S_{mt}$ fills the place where the data is masked, i.e., $M_{i,k}=1$.
\begin{equation}
    \mathbf{S^{Ti}_i}=TSLayer(\{\mathbf{S_{i,0}^{Ti}}, \mathbf{S_{mt}^{Ti}},\cdots, \mathbf{S_{i,P}^{Ti}}\})
\end{equation}
\begin{equation}
    \mathbf{S^{Am}_i}=TSLayer(\{\mathbf{S_{i,0}^{Am}}, \mathbf{S_{mt}^{Am}},\cdots, \mathbf{S_{i,P}^{Am}}\})
\end{equation}
\begin{equation}
    \mathbf{S^{Ph}_i}=TSLayer(\{\mathbf{S_{i,0}^{Ph}}, \mathbf{S_{mt}^{Ph}},\cdots, \mathbf{S_{i,P}^{Ph}}\})
\end{equation}
Here, $\mathbf{S_{0,i}^{Ti}}$ indicates the patch data of node $i$ at time 0 and $\mathbf{S^{Ti}}$ indicates the patch series in the time domain, and $\mathbf{S^{Am}}$, $\mathbf{S^{Ph}}$ indicates amplitude and phase domain respectively.
Next, the concatenated data of the time, amplitude, and phase domains is then fed to a Cross-Domain Aggregator to aggregate the information of each domain.
We use a Transformer layer ($TSLayer(\cdot)$) as the Cross-Domain Aggregator here.
\begin{equation}
    \mathbf{H^{Ti}_i}, \mathbf{H^{Am}_i}, \mathbf{H^{Ph}_i} = TSLayer(Concat\{\mathbf{S^{Ti}_i}, \mathbf{S^{Am}_i}, \mathbf{S^{Ph}_i}\})
\end{equation}
After the information of the three domains is aggregated, we expect the model to further integrate the knowledge of the neighboring nodes. 
To achieve this, a Graph Neural Network ($GNN(\cdot)$) encoder is incorporated into the architecture as a Cross-Space Aggregator, dedicated to aggregating information for each domain separately.
\begin{equation}
    \mathbf{H^{Ti}_i}=GNN\{\mathbf{H^{Ti}_j}|j\in\mathcal{N}_i\}
\end{equation}
\begin{equation}
    \mathbf{H^{Am}_i}=GNN\{\mathbf{H^{Am}_j}|j\in\mathcal{N}_i\}
\end{equation}
\begin{equation}
    \mathbf{H^{Ph}_i}=GNN\{\mathbf{H^{Ph}_j}|j\in\mathcal{N}_i\}
\end{equation}
Here, $\mathcal{N}_i$ denotes the set of neighboring nodes of node $i$ in the raw graph $\mathbf{A}$. The GNN layer is applied independently to each domain, enabling the model to capture the contextual information from neighboring nodes in the respective time, amplitude, and phase domains.
Finally, we feed the data into another Cross-Domain Aggregator again.
\begin{equation}
    \mathbf{\hat{H}^{Ti}_i}, \mathbf{\hat{H}^{Am}_i}, \mathbf{\hat{H}^{Ph}_i} = TSLayer(Concat\{\mathbf{H^{Ti}_i}, \mathbf{H^{Am}_i}, \mathbf{H^{Ph}_i}\})
\end{equation}
Here $\mathbf{\hat{H}^{Ti}_i}$, $\mathbf{\hat{H}^{Am}_i}$, $\mathbf{\hat{H}^{Ph}_i}$indicates the output embedding of Cross-Domain Spatial-Temporal Encoder in the time, amplitude, phase domains respectively.

\noindent\textbf{Reconstruction Module}
Once we obtain the output embeddings from the encoder, we can reconstruct the masked input patches. This reconstruction process involves applying linear transformations to the output embeddings.
\begin{equation}
\mathbf{\hat{S}^{Ti}_{i,j}}=Linear(\mathbf{\hat{H}^{Ti}_{i,j}})
\end{equation}
\begin{equation}
\mathbf{\hat{S}^{Am}_{i,j}}=Linear(\mathbf{\hat{H}^{Am}_{i,j}})
\end{equation}
\begin{equation}
\mathbf{\hat{S}^{Ph}_{i,j}}=Linear(\mathbf{\hat{H}^{Ph}_{i,j}})
\end{equation}
Then, the reconstruction loss is the Mean Square Error (MSE) of the masked patches.
\begin{multline}
    \mathcal{L}_{re}=\sum_{i,j}\mathbf{M^{Ti}_{ij}}(\mathbf{\hat{S}^{Ti}_{i,j}}-\mathbf{S^{Ti}_{i,j}})^2+\sum_{i,j}\mathbf{M^{Am}_{ij}}(\mathbf{\hat{S}^{Am}_{i,j}}-\mathbf{S^{Am}_{i,j}})^2+\sum_{i,j}\mathbf{M^{Ph}_{ij}}(\mathbf{\hat{S}^{Ph}_{i,j}}-\mathbf{S^{Ph}_{i,j}})^2
\end{multline}
By optimizing the encoder to minimize the reconstruction loss, we encourage the model to learn representations that capture the essential information needed to reconstruct the masked patches accurately.

\noindent\textbf{Contrastive Module}
To further improve the representation learning and enhance the expressiveness of the learned embedding space, a contrastive loss is introduced.
We introduce a novel approach to constructing augmented data that effectively utilizes the frequency domain information.
Recognizing the significant informational content present in the amplitude domain of traffic data, we randomly switch the amplitude domain within a given batch of data, thereby producing augmented samples.
\begin{equation}
    \mathbf{\widetilde{S}^{Ti}_i}=\mathbf{S^{Ti}_i}, \mathbf{\widetilde{S}^{Ph}_i}=\mathbf{S^{Ph}_i}
\end{equation}
\begin{equation}
    \mathbf{\widetilde{S}^{Am}_i}=\mathbf{S^{Am}_j}\ \ i,j\in SameBatch
\end{equation}
Here $\mathbf{\widetilde{S}^{Ti}_i}, \mathbf{\widetilde{S}^{Ph}_i}, \mathbf{\widetilde{S}^{Am}_i}$ is the augmented data in the three domains.
Then, we mask the data and feed the data into the Cross-Domain Spatial-Temporal Encoder $CDEnc(\cdot)$ to get the augmented embedding.
\begin{equation}
    \mathbf{\widetilde{H}^{Ti}_i}, \mathbf{\widetilde{H}^{Am}_i}, \mathbf{\widetilde{H}^{Ph}_i} = CDEnc(\mathbf{\widetilde{S}^{Ti}_i}, \mathbf{\widetilde{S}^{Am}_i}, \mathbf{\widetilde{S}^{Ph}_i})
\end{equation}
Subsequently, a linear layer is introduced as a Domain Embedding Aggregator, responsible for merging the embeddings from the three domains.
\begin{equation}
    \mathbf{\widetilde{H}_i}=Linear(Concat\{\mathbf{\widetilde{H}^{Ti}_i},\mathbf{\widetilde{H}^{Am}_i},\mathbf{\widetilde{H}^{Ph}_i}\})
\end{equation}
Similarly, the embedding of the original data $\mathbf{\hat{H}_i}$ is generated.
\begin{equation}
    \mathbf{\hat{H}_i}=Linear(Concat\{\mathbf{\hat{H}^{Ti}_i},\mathbf{\hat{H}^{Am}_i},\mathbf{\hat{H}^{Ph}_i}\})
\end{equation}
Finally, we want the data sample and its augmented pair to have similar embeddings.
By doing this, the encoder is more robust and could capture the essential dynamics of the frequency domain of the data.
We employ the NT-Xent Loss~\cite{chen2020simple} here to optimize the encoder.
\begin{equation}
    \mathcal{L}_{con}=-\sum_i^N\log\frac{e^{sim(\mathbf{\hat{H}_i},\mathbf{\widetilde{H}_i})}}{\sum_j e^{sim(\mathbf{\hat{H}_i}, \mathbf{\widetilde{H}_j})}}
\end{equation}
Here $sim$ is the cosine similarity and the negative samples $j$ are randomly chosen 10\% from the $N$ nodes. 

The total loss of the pre-training stage is the sum of the two losses and it aims to update the Cross-Domain Spatial-Temporal Encoder. 
\begin{equation}
    \mathcal{L}=\mathcal{L}_{re}+\alpha \mathcal{L}_{con}
\end{equation}

\subsection{Fine-Tuning}
In the fine-tuning stage, we aim to utilize the pre-trained Cross-Domain Spatial-Temporal Encoder to help forecast the traffic with few-shot training data in the target city. 
Firstly, we employ a masking and reconstructing technique to enrich the few-shot training data.
Next, we utilize the knowledge learned by the pre-trained encoder to improve the graph structure representation of the target city.
Finally, we apply a spatial-temporal short-term model to predict future traffic

\noindent\textbf{Training Data Enriching:} 
\begin{figure}[!t]
    \centering
    \includegraphics[width=0.7  \linewidth]{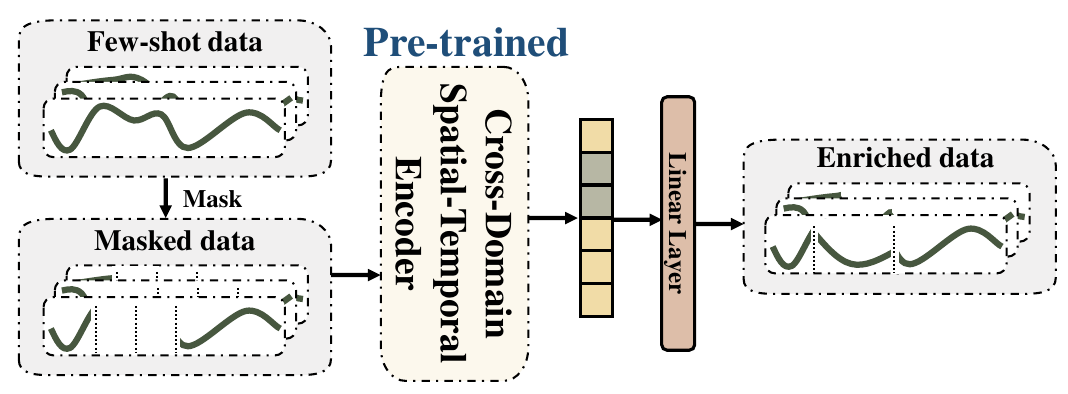}
    \caption{Illustration of Training Data Enriching Module.}
    \label{fig:method_reconst}
\end{figure}
The target city may only have very little data.
Directly utilizing the few-shot data could lead to over-fitting and lack of expressiveness of the fine-tuned model.
Consequently, we use the pre-trained Cross-Domain Spatial-Temporal Encoder to augment the training data to create more robust training samples.
As shown in Fig.~\ref{fig:method_reconst}, we first mask some part of the training data and use the masked token to replace the masked part.
Then, we use the pre-trained encoder to reconstruct the masked data.
The reconstructed data is combined with the unmasked data to form the augmented training set. 
This approach addresses the limited availability of data of the target city, mitigates the risk of over-fitting, and enhances the expressiveness of the fine-tuned model.

\noindent{\textbf{Momentum Graph:}}
Due to the potential presence of diverse sources of noise and uncertainties during the generation of the initial traffic graph structure, we exploit the learned knowledge from the pre-trained encoder to enhance the graph structure of the target city.
Specifically, the input data sampled from the enriched training dataset is firstly Fourier Transformed.
Then, the data of the three domains is fed into the pre-trained Cross-Domain Spatial-Temporal Encoder.
\begin{equation}
    \mathbf{\hat{H}^{Ti}_i}, \mathbf{\hat{H}^{Am}_i}, \mathbf{\hat{H}^{Ph}_i} = CDEnc(\mathbf{{S}^{Ti}_i}, \mathbf{{S}^{Am}_i}, \mathbf{{S}^{Ph}_i})
\end{equation}
Subsequently, the embedding of three domains is aggregated.
\begin{equation}
\mathbf{\hat{H}_i}=\mathbf{\hat{H}^{Ti}_i}+\mathbf{\hat{H}^{Am}_i}+\mathbf{\hat{H}^{Ph}_i}
\end{equation}
The output embedding $\mathbf{\hat{H}}\in\mathbb{R}^{N\times D}$ generates the meta-graph $\mathbf{\hat{A}_{meta}}$ via inner product.
\begin{equation}
    \mathbf{\hat{A}_{meta}}=SoftMax(\mathbf{\hat{H}}\mathbf{\hat{H}}^T)
\end{equation}
In order to enhance the resilience of the learned graph structure and prevent it from over-fitting the limited training data, we utilize the momentum-updating technique to preserve the refined graph structure of the target city, denoted as $\mathbf{\hat{A}}$. 
Formally, in the $k$-th graph reconstruction, the graph structure is updated as follows.
\begin{equation}
    \mathbf{\hat{A}_k} = \tau \times \mathbf{\hat{A}_{meta}} + (1 - \tau) \times \mathbf{\hat{A}_{k-1}}
\end{equation}
\begin{equation}
    \mathbf{\hat{A}_0} = \mathbf{A}
\end{equation}

\noindent{\textbf{Forecasting:}} After generating the graph structure $\mathbf{\hat{A}_k}$, it is subsequently utilized as input to a downstream Spatial-Temporal model (STmodel), together with the short-term raw data. In this context, we use Graph Wavenet~\cite{wu2019graph} as the backbone STmodel framework.
\begin{equation}
    \mathbf{\hat{H}^{ST}}=STmodel(\mathbf{S^{Ti}_{P}},\mathbf{\hat{A}_k})
\end{equation}
Finally, we concatenate the embedding of the STmodel and the pre-trained encoder to forecast future traffic.
\begin{equation}
    \mathcal{\hat{Y}}=Linear(\mathbf{\hat{H}}||\mathbf{\hat{H}^{ST}})
\end{equation}
The Mean Square Loss is utilized to optimize the STmodel and Linear layers.
\begin{equation}
    \mathcal{L}=\frac{1}{NT_fC}\sum_{i=1}^N\sum_{j=1}^{T_f}\sum_{k=1}^{C}(\mathcal{Y}_{ijk}-\hat{\mathcal{Y}}_{ijk})^2
\end{equation}
\begin{table}[tp]
    \caption{Statistical details of traffic datasets.}
    \label{tab:data}
    \centering
    \resizebox{0.63\linewidth}{!}{
    \begin{tabular}{c|c|c|c|c}
    \toprule
         & PEMS-BAY & METR-LA & Chengdu & Shenzhen\\
         \midrule
         \# of Nodes & 325 & 207 & 524 & 627\\ 
         \# of Edges & 2,694 & 1,722 & 1,120 & 4,845\\
         Interval & 5 min & 5 min & 10 min & 10 min\\
         \# of Time Step & 52,116 & 34,272 & 17,280 & 17,280\\
         Mean & 61.7768 & 58.2749 & 29.0235 & 31.0092\\
         Std & 9.2852 & 13.1280 & 9.6620 & 10.9694\\ 
    \bottomrule
    \end{tabular}
    }
\end{table}

\section{Experiment}
This section presents a thorough evaluation of our proposed framework FEPCross. Specifically, We would address the following research questions.
\begin{itemize}
    \item How does FEPCross perform compared to other baselines in the task of few-shot traffic forecasting?
    \item Does each component contribute to the final result?
    \item Do the proposed modules effectively enhance the expressiveness of the pre-trained encoder?
    \item Does the Cross-Domain Spatial-Temporal Encoder successfully fuse the information of three domains?
\end{itemize}


\subsection{Experiment Settings}

\noindent{\textbf{Dataset:}}
We evaluate our proposed framework on four real-world public datasets: \textit{PEMS-BAY}, \textit{METR-LA}~\cite{li2017diffusion}, \textit{Chengdu}, \textit{Shenzhen}~\cite{Didi}.
These datasets contain months of traffic speed data and the details of these data are listed in Table~\ref{tab:data}.

\noindent{\textbf{Few-shot Setting:}}
Following the experimental setting of previous research~\cite{liu2023cross,lu2022spatio,liu2024multi}, we choose two cities and set one as the source city and the other one as the target city. 
We use the full data of the source city as the pre-training data, the two-day few-shot data of the target city as the fine-tuning data, and the rest of the data of the target city as the testing data.

\noindent{\textbf{Details:}}
In the pre-training stage, we use $T_h=288$ and $P=24$, which means one-day data is divided into 24 patches as the input.
The loss weight $\alpha$ is set to 1.
The mask ratio of Pre-training is set to 75\% according to~\cite{liu2023cross,shao2022pre}.
The learning rate of Pre-training is set to 0.0001.
In the Forecasting module, we use $T_f=12$ to forecast the future 12 steps of data.
The momentum ratio $\tau$ is set to 0.1.
The Adam optimizer has a learning rate of 0.001 and a weight decay of 0.01.
The dimension of $\mathbf{H}$ is set to 128.
The experiment is implemented by Pytorch 1.10.0 on RTX3090.
We evaluate the performances of Mean Absolute Error (MAE) and Mean Absolute Percentage Error (MAPE).
Code and data are released in~\url{https://github.com/zhyliu00/FEPCross}.
\begin{equation}
    MAPE=\frac{1}{s}\sum_{i=1}^s|\frac{y_i-\hat{y_i}}{y_i}|,\ \ MAE=\frac{1}{s}\sum_{i=1}^s|{y_i-\hat{y_i}}|
\end{equation}

\section{RQ1: Overall Performance}
We select 13 baselines, including traditional statistic methods, deep traffic forecasting methods, time series forecasting methods, and cross-city traffic forecasting methods, to evaluate the effectiveness of FEPCross on the few-shot forecasting task. 
To guarantee the fairness of the comparison, the deep traffic forecasting methods and time series forecasting methods are firstly trained by Reptile~\cite{nichol2018first} meta-learning framework on the data of source city and then fine-tuned on the few-shot data on target city.
For each method, we keep the model architecture the same as the original paper and search the hyper-parameters in the same log space while fixing other parameters (for example, we search the fine-tune learning rate in [1e-5, 1e-4, 1e-3, 1e-2, 1e-1] for all methods). 
\begin{itemize}[leftmargin=*]
\item Traditional methods: 
    \textbf{HA} uses the average of previous periods as predictions. \textbf{SVR} uses Support Vector Machines to do regression tasks.
\item Deep traffic forecasting methods: 
    \textbf{DCRNN}~\cite{li2017diffusion} uses diffusion techniques. \textbf{GWN}~\cite{wu2019graph} utilizes adaptive graph and dilated causal convolution. \textbf{STFGNN}~\cite{li2021spatial} uses DTW distance to construct the temporal graph. \textbf{DSTAGNN}~\cite{lan2022dstagnn} constructs spatial-temporal aware graph. \textbf{FOGS}~\cite{rao2022fogs} predicts the first-order difference.
\item Time series forecasting methods: 
    \textbf{DLinear}~\cite{zeng2023transformers} uses linear layers and \textbf{FEDFormer}~\cite{zhou2022fedformer} utilized a frequency-enhanced Transformer while they both have decomposition module.  
\item Cross-city traffic forecasting methods: 
    \textbf{ST-GFSL}~\cite{lu2022spatio} learns the metaknowledge to generate the parameter of the models. 
    \textbf{STEP}~\cite{shao2022pre} reconstructs the masked traffic to do per-taining on source datasets. 
    \textbf{TransGTR}~\cite{jin2023transferable} build transferable graph structure across cities.
    \textbf{TPB}~\cite{liu2023cross} constructs the traffic pattern bank from multiple source cities.
\end{itemize}

The performance evaluation results are presented in Table~\ref{tab:performance}.
Based on the information provided in the table, the following observations can be made.
(1) FEPCross surpasses the baseline methods in both short-term and long-term traffic forecasting tasks, exhibiting a notable average improvement of 15.84\% in Mean Absolute Percentage Error (MAPE) and 16.77\% in Mean Absolute Error (MAE).
(2) The cross-city traffic forecasting methods exhibit superior performance compared to the other three categories of baseline methods. 
This outcome underscores the significance of explicitly addressing the cross-city few-shot traffic forecasting problem, emphasizing the need for dedicated modeling approaches.
(3) Among the cross-city traffic forecasting methods, FEPCross performs the best, which demonstrates that FEPCross captures resilient and generalizable frequency-enhanced traffic information across cities.
(4) Notably, FEDFormer has high errors in short-term forecasting, which is attributed to that it is primarily tailored for long-term time series forecasting tasks.

\begin{table*}[!t]
\caption{Overall performance of few-shot traffic forecasting on \textit{PEMS-BAY}, \textit{METR-LA}, \textit{Chengdu}, and \textit{Shenzhen}. Metr-LA$\rightarrow$PEMS-BAY means the source city is METR-LA and the target city is PEMS-BAY. The mean and standard deviation of the results in 5 runs is shown. In each column, the best result is highlighted in bold and grey, and the second-best result is underlined. Marker * and ** indicates the mean of the results is statistically significant  (* means t-test with p-value $<$ 0.05 and ** means t-test with p-value $<$ 0.01).}
\centering

\label{tab:performance}
\resizebox{0.94\linewidth}{!}{

\begin{tabular}{c|c c c c c c||c c c c c c}
\toprule
& \multicolumn{6}{c||}{\textbf{METR-LA$\rightarrow$PEMS-BAY}} & \multicolumn{6}{c}{\textbf{PEMS-BAY$\rightarrow$METR-LA}}\\
\cline{2-7}
\cline{8-13}
& \multicolumn{2}{c}{5 min} & \multicolumn{2}{c}{15 min} & \multicolumn{2}{c||}{30 min}& \multicolumn{2}{c}{5 min} & \multicolumn{2}{c}{15 min} & \multicolumn{2}{c}{30 min}\\
\cline{2-7}
\cline{8-13}
 &MAPE(\%) & MAE & MAPE(\%) & MAE & MAPE(\%)  & MAE & MAPE(\%) & MAE & MAPE(\%) & MAE & MAPE(\%)  & MAE\\ 

\midrule
\midrule
HA & 9.33 & 3.99 & 9.33 & 3.99  & 9.33 & 3.99 & 14.15 & 5.21 &  14.15 & 5.21 & 14.15 & 5.21 
\\ 
SVR&  8.07 & 3.79 & 7.96 & 3.70 & 7.95 & 3.71 & 14.02 & 4.97  &14.59 & 5.11 & 14.02& 5.04
 \\ 
\midrule
DCRNN & 3.66$\pm$0.21 & 1.64$\pm$0.04 & 4.45$\pm$0.21 & 2.04$\pm$0.06 & 5.80$\pm$0.15 & 2.66$\pm$0.06 & 7.88$\pm$0.21 & 3.13$\pm$0.14 & 9.43$\pm$0.25 & 3.56$\pm$0.11 & 11.91$\pm$0.42 & 4.31$\pm$0.18

 \\ 
GWN & 3.67$\pm$0.70 & 1.49$\pm$0.18 & 5.14$\pm$0.76 & 2.07$\pm$0.16 & 7.04$\pm$1.04 & 2.72$\pm$0.23 & 6.67$\pm$0.48 & 2.78$\pm$0.08 & 9.83$\pm$0.80 & 3.49$\pm$0.15 & 12.77$\pm$1.07 & 4.36$\pm$0.21

 \\ 
STFGNN & 3.21$\pm$0.50 & 1.49$\pm$0.11 & 4.05$\pm$0.17 & 1.93$\pm$0.06 & 5.68$\pm$0.18 & 2.66$\pm$0.13 & 7.57$\pm$0.46 & 3.40$\pm$0.32 & 9.78$\pm$0.42 & 3.93$\pm$0.20 & 12.50$\pm$0.65 & 4.52$\pm$0.25

  \\ 
DSTAGNN & 3.36$\pm$0.49 & 1.67$\pm$0.32 & 4.80$\pm$0.80 & 2.34$\pm$0.54 & 6.29$\pm$0.66 & 2.85$\pm$0.37 & 7.11$\pm$0.57 & 2.82$\pm$0.14 & 10.18$\pm$1.23 & 3.59$\pm$0.47 & 12.56$\pm$1.11 & 4.42$\pm$0.37

  \\ 
FOGS & 2.87$\pm$0.28 & 1.38$\pm$0.08 & 3.94$\pm$0.24 & 1.87$\pm$0.05 & 5.64$\pm$0.30 & 2.62$\pm$0.060 & 7.20$\pm$0.43 & 3.23$\pm$0.29 & 9.56$\pm$0.55 & 3.84$\pm$0.25 & 12.41$\pm$0.63 & 4.48$\pm$0.25
  \\ 
\midrule
FEDFormer & 6.01$\pm$0.55 & 3.08$\pm$0.37 & 5.87$\pm$0.51 & 2.89$\pm$0.24 & 5.65$\pm$0.46 & 2.69$\pm$0.24 & 14.35$\pm$0.92 & 5.02$\pm$0.40 & 13.12$\pm$1.00 & 4.98$\pm$0.35 & 12.24$\pm$0.34 & 4.79$\pm$0.40
\\
DLinear & 2.64$\pm$0.21 & 1.44$\pm$0.14 & 4.04$\pm$0.14 & 2.09$\pm$0.07 & 5.64$\pm$0.46 & 2.80$\pm$0.24 & 6.63$\pm$0.31 & 2.84$\pm$0.06 & 9.34$\pm$0.69 & 3.66$\pm$0.22 & 11.87$\pm$0.42 & 4.47$\pm$0.10
\\
\midrule
ST-GFSL & 2.88$\pm$0.14 & 1.38$\pm$0.04 & 4.12$\pm$0.08 & 1.99$\pm$0.05 & 5.46$\pm$0.15 & 2.48$\pm$0.09 & 7.01$\pm$0.05 & 2.80$\pm$0.03 & 9.17$\pm$0.11 & 3.43$\pm$0.08 & 13.47$\pm$0.68 & 4.45$\pm$0.16
\\ 
STEP & \underline{2.39$\pm$0.18} & \underline{1.17$\pm$0.00} & \underline{3.70$\pm$0.19} & 1.75$\pm$0.02 & \underline{5.29$\pm$0.16} & 2.36$\pm$0.04 & 7.23$\pm$1.18 & \underline{2.74$\pm$0.13} & 9.32$\pm$0.34 & 3.32$\pm$0.01 & 11.89$\pm$0.54 & 4.15$\pm$0.26
\\ 

TransGTR & 3.62$\pm$0.14 & 1.47$\pm$0.03 & 4.51$\pm$0.17 & 1.89$\pm$0.02 & 5.57$\pm$0.20 & \underline{2.35$\pm$0.04} & 7.73$\pm$0.08 & 2.86$\pm$0.08 & 8.99$\pm$0.12 & 3.31$\pm$0.03 & \underline{11.51$\pm$0.61} & 4.07$\pm$0.04
\\ 

TPB & 2.55$\pm$0.29 & 1.19$\pm$0.08 & 3.86$\pm$0.29 & \cellcolor{lgray}{\textbf{1.70$\pm$0.07}} & 5.43$\pm$0.31 & 2.36$\pm$0.06 & \underline{6.61$\pm$0.11} & 2.78$\pm$0.16 & \underline{8.97$\pm$0.64} & \underline{3.29$\pm$0.10} & 11.97$\pm$1.11 & \cellcolor{lgray}{\textbf{3.98$\pm$0.09}}
\\ 

\midrule
FEPCross & \cellcolor{lgray}{\textbf{1.98$\pm$0.02**}} & \cellcolor{lgray}{\textbf{1.04$\pm$0.01**}} & \cellcolor{lgray}{\textbf{3.48$\pm$0.04**}} & \underline{1.71$\pm$0.01} & \cellcolor{lgray}{\textbf{5.04$\pm$0.07**}} & \cellcolor{lgray}{\textbf{2.31$\pm$0.02**}} & \cellcolor{lgray}{\textbf{5.90$\pm$0.04*}} & \cellcolor{lgray}{\textbf{2.50$\pm$0.07*}} & \cellcolor{lgray}{\textbf{8.52$\pm$0.26*}} & \cellcolor{lgray}{\textbf{3.23$\pm$0.01*}} & \cellcolor{lgray}{\textbf{11.16$\pm$0.20**}} & \cellcolor{lgray}{\textbf{3.98$\pm$0.01}}
\\ 

\midrule
\midrule
& \multicolumn{6}{c||}{\textbf{Shenzhen$\rightarrow$Chengdu}} & \multicolumn{6}{c}{\textbf{Chengdu$\rightarrow$Shenzhen}}\\
\cline{2-7}
\cline{8-13}
& \multicolumn{2}{c}{10 min} & \multicolumn{2}{c}{30 min} & \multicolumn{2}{c||}{60 min}& \multicolumn{2}{c}{10 min} & \multicolumn{2}{c}{30 min} & \multicolumn{2}{c}{60 min}\\
\cline{2-7}
\cline{8-13}
 &MAPE(\%) & MAE & MAPE(\%) & MAE & MAPE(\%)  & MAE & MAPE(\%) & MAE & MAPE(\%) & MAE & MAPE(\%)  & MAE\\ 
\midrule
\midrule
HA & 16.27 & 3.77 & 16.27 & 3.77  & 16.27 & 3.77 & 17.33 & 4.43 & 17.33 & 4.43 & 17.33 & 4.43 
\\ 
SVR & 15.39 & 3.58  & 15.57 & 3.62  & 15.79 & 3.71 & 15.01 & 4.02 & 15.68  & 4.17 & 16.12 & 4.35
 \\ 
\midrule
DCRNN & 10.38$\pm$0.24 & 2.55$\pm$0.02 & 14.07$\pm$0.31 & 3.23$\pm$0.03 & 15.26$\pm$0.23 & 3.55$\pm$0.04 & 8.26$\pm$0.15 & 2.19$\pm$0.05 & 10.66$\pm$0.09 & 2.70$\pm$0.04 & 11.39$\pm$0.21 & 2.95$\pm$0.06

 \\ 
GWN & 10.49$\pm$0.63 & 2.57$\pm$0.07 & 14.74$\pm$0.99 & 3.36$\pm$0.11 & 15.82$\pm$0.83 & 3.70$\pm$0.06 & 8.62$\pm$0.29 & 2.25$\pm$0.04 & 11.46$\pm$0.44 & 2.88$\pm$0.08 & 12.21$\pm$0.55 & 3.15$\pm$0.11

  \\ 
STFGNN & 11.79$\pm$0.86 & 2.90$\pm$0.11 & 15.78$\pm$0.98 & 3.59$\pm$0.09 & 16.95$\pm$0.85& 3.95$\pm$0.07 & 9.96$\pm$0.54 & 2.55$\pm$0.08 & 12.60$\pm$0.81 & 3.09$\pm$0.10 & 13.21$\pm$0.74 & 3.35$\pm$0.09

  \\ 
DSTAGNN & 10.37$\pm$0.39 & 2.55$\pm$0.04 & 15.31$\pm$1.15 & 3.46$\pm$0.31 & 15.47$\pm$0.88 & 3.62$\pm$0.10 & 9.27$\pm$ 1.00 & 2.48$\pm$0.32 & 11.94$\pm$0.80 & 2.94$\pm$0.15 & 13.17$\pm$1.18 & 3.41$\pm$0.43

  \\ 
FOGS & 11.63$\pm$0.82 & 2.83$\pm$0.10 & 15.79$\pm$0.99 & 3.56$\pm$0.11 & 16.89$\pm$0.99 & 3.91$\pm$0.10 & 9.32$\pm$0.13 & 2.44$\pm$0.02 & 12.00$\pm$0.200 & 3.01$\pm$0.03 & 12.64$\pm$0.18 & 3.28$\pm$0.05
  \\ 
\midrule
FEDFormer & 13.99$\pm$0.76 & 3.26$\pm$0.17 & 13.53$\pm$0.86 & 3.06$\pm$0.17 & 13.49$\pm$0.62 & 3.17$\pm$0.19 & 11.88$\pm$0.87 & 2.87$\pm$0.29 & 11.55$\pm$0.70 & 2.76$\pm$0.31 & 11.65$\pm$0.35 & 2.78$\pm$0.23
\\
DLinear & 10.76$\pm$0.08 & 2.72$\pm$0.02 & 14.15$\pm$0.27 & 3.40$\pm$0.07 & 13.98$\pm$0.32 & 3.45$\pm$0.09 & 8.79$\pm$0.06 & 2.35$\pm$0.01 & 11.07$\pm$0.08 & 2.85$\pm$0.04 & 11.20$\pm$0.12 & 3.00$\pm$0.05
\\
\midrule
ST-GFSL & 9.73$\pm$0.48 & 2.40$\pm$0.06 & 12.91$\pm$0.88 & 2.94$\pm$0.11 & 14.26$\pm$1.03 & 3.35$\pm$0.11 & 8.24$\pm$0.55 & 2.04$\pm$0.03 & 10.67$\pm$0.89 & 2.46$\pm$0.07 & 11.91$\pm$1.00 & 2.78$\pm$0.10
\\ 

STEP & 9.21$\pm$0.12 & 2.31$\pm$0.03 & \underline{12.06$\pm$0.01} & 2.88$\pm$0.04 & 13.30$\pm$0.11 & 3.17$\pm$0.04 & 7.62$\pm$0.09 & 1.95$\pm$0.00 & 9.92$\pm$0.18 & 2.44$\pm$0.02 & 10.88$\pm$0.14 & 2.72$\pm$0.07
\\ 

TransGTR & 9.51$\pm$0.13 & 2.28$\pm$0.04 & 12.24$\pm$0.18 & 2.79$\pm$0.02 & 13.38$\pm$0.17 & 3.02$\pm$0.03 & 7.94$\pm$0.09 & 2.04$\pm$0.04 & \underline{9.83$\pm$0.08} & 2.46$\pm$0.04 & 10.60$\pm$0.04 & \underline{2.64$\pm$0.03}
\\ 

TPB & \underline{9.18$\pm$0.13} & \underline{2.19$\pm$0.02} & 12.10$\pm$0.19 & \underline{2.75$\pm$0.04} & \underline{13.21$\pm$0.27} & \underline{3.02$\pm$0.04} & \underline{7.61$\pm$0.13} & \underline{1.94$\pm$0.07} & 9.97$\pm$0.16 & \underline{2.38$\pm$0.08} & \underline{10.57$\pm$0.20} & 2.68$\pm$0.08
\\ 

\midrule
FEPCross & \cellcolor{lgray}{\textbf{8.94$\pm$0.09**}} & \cellcolor{lgray}{\textbf{2.16$\pm$0.01**}} & \cellcolor{lgray}{\textbf{11.42$\pm$0.12**}} & \cellcolor{lgray}{\textbf{2.63$\pm$0.02**}} & \cellcolor{lgray}{\textbf{12.04$\pm$0.05**}} & \cellcolor{lgray}{\textbf{2.78$\pm$0.01**}} & \cellcolor{lgray}{\textbf{7.35$\pm$0.05**}} & \cellcolor{lgray}{\textbf{1.89$\pm$0.03**}} & \cellcolor{lgray}{\textbf{9.26$\pm$0.08**}} & \cellcolor{lgray}{\textbf{2.32$\pm$0.04**}} & \cellcolor{lgray}{\textbf{9.81$\pm$0.07**}} & \cellcolor{lgray}{\textbf{2.46$\pm$0.03**}}
\\ 
\bottomrule
\bottomrule

\end{tabular}
}
\end{table*}


\section{RQ2: Ablation Study}

\begin{table*}[!t]
\caption{Ablation Study. The mean and standard deviation of the results on the target city in 5 runs are shown. In each column, the best result is highlighted in bold and grey. P-T and F-T mean the modules proposed in the pre-training stage and the fine-tuning stage respectively.}
\centering
\label{tab:ablation}
\resizebox{0.94\linewidth}{!}{

\begin{tabular}{c|c|c c c c c c||c c c c c c}
\toprule
\multicolumn{2}{c|}{} & \multicolumn{6}{c||}{\textbf{PEMS-BAY$\rightarrow$METR-LA}} & \multicolumn{6}{c}{\textbf{Shenzhen$\rightarrow$Chengdu}}\\
\cline{3-8}
\cline{9-14}
\multicolumn{2}{c|}{} & 5 min& 15 min & 30 min  & 5 min& 15 min & 30 min & 10 min & 30 min & 60 min & 10 min & 30 min & 60 min\\

\cline{3-8}
\cline{9-14}
\multicolumn{2}{c|}{} & MAE & MAE & MAE & MAE & MAE & MAE & MAE & MAE & MAE & MAE & MAE & MAE \\

\midrule
\midrule
\multirow{4}{*}{\rotatebox{90}{P-T}}& Pretrain base & 7.23$\pm$1.18 & 2.74$\pm$0.13 &9.32$\pm$0.34 &3.32$\pm$0.01  &11.89$\pm$0.54 &4.15$\pm$0.26&
9.21$\pm$0.12 & 2.31$\pm$0.03 & 12.06$\pm$0.01 &2.88$\pm$0.04 &13.30$\pm$0.11  &3.17$\pm$0.04\\
 
& Pretrain base+F & 6.35$\pm$0.18 & 2.75$\pm$0.09 & 8.91$\pm$0.20 & 3.47$\pm$0.07 & 11.78$\pm$0.24 & 4.43$\pm$0.12& 9.31$\pm$0.13 & 2.24$\pm$0.02 & 11.87$\pm$0.10 & 2.75$\pm$0.02 & 12.85$\pm$0.13 & 2.96$\pm$0.03  \\

& Pretrain base+F+D & 6.25$\pm$0.07 & 2.64$\pm$0.06 & 8.73$\pm$0.11 & 3.38$\pm$0.08 & 11.36$\pm$0.16 & 4.27$\pm$0.06 & 9.19$\pm$0.05 & 2.22$\pm$0.02 & 11.64$\pm$0.10 & 2.69$\pm$0.02 & 12.39$\pm$0.09 & 2.86$\pm$0.02\\ 

& Pretrain base+F+D+S & 6.09$\pm$0.19 & 2.57$\pm$0.05 & 8.54$\pm$0.38 & 3.31$\pm$0.06 & 11.31$\pm$0.46 & 4.25$\pm$0.03 &  9.08$\pm$0.06 & 2.19$\pm$0.01 & 11.65$\pm$0.12 & 2.67$\pm$0.02 & 12.49$\pm$0.11 & 2.83$\pm$0.03\\ 

\midrule

\multirow{2}{*}{\rotatebox{90}{F-T}}& Finetune Base & 6.13$\pm$0.15 & 2.58$\pm$0.03 & 8.80$\pm$0.38 & 3.34$\pm$0.06 & 11.49$\pm$0.48 & 4.24$\pm$0.13 & 9.44$\pm$0.04 & 2.33$\pm$0.03 & 12.22$\pm$0.14 & 2.92$\pm$0.09 & 12.67$\pm$0.14 & 3.01$\pm$0.09
  \\ 
& Finetune Base+M & 6.11$\pm$0.21 & 2.54$\pm$0.15 & 8.60$\pm$0.23 & 3.26$\pm$0.19 & 11.18$\pm$0.27 & 4.09$\pm$0.23 & \cellcolor{lgray}{\textbf{8.89$\pm$0.07}} & \cellcolor{lgray}{\textbf{2.16$\pm$0.01}} & 11.55$\pm$0.13 & 2.66$\pm$0.01 & 12.15$\pm$0.04 & 2.78$\pm$0.02

\\

\midrule
& FEPCross&  \cellcolor{lgray}{\textbf{5.90$\pm$0.04}} &  \cellcolor{lgray}{\textbf{2.50$\pm$0.07}} &  \cellcolor{lgray}{\textbf{8.52$\pm$0.26}} &  \cellcolor{lgray}{\textbf{3.23$\pm$0.01}} &  \cellcolor{lgray}{\textbf{11.16$\pm$0.25}} &  \cellcolor{lgray}{\textbf{3.98$\pm$0.01}} &  {8.94$\pm$0.09} &  \cellcolor{lgray}{\textbf{2.16$\pm$0.01}} &  \cellcolor{lgray}{\textbf{11.42$\pm$0.12}} &  \cellcolor{lgray}{{\textbf{2.63$\pm$0.02}}} &  \cellcolor{lgray}{\textbf{12.04$\pm$0.05}} &  \cellcolor{lgray}{\textbf{2.77$\pm$0.01}} \\ 


\bottomrule
\bottomrule

\end{tabular}
}
\end{table*}

\begin{figure}[!t]
    \centering
    \includegraphics[width=0.74\linewidth]{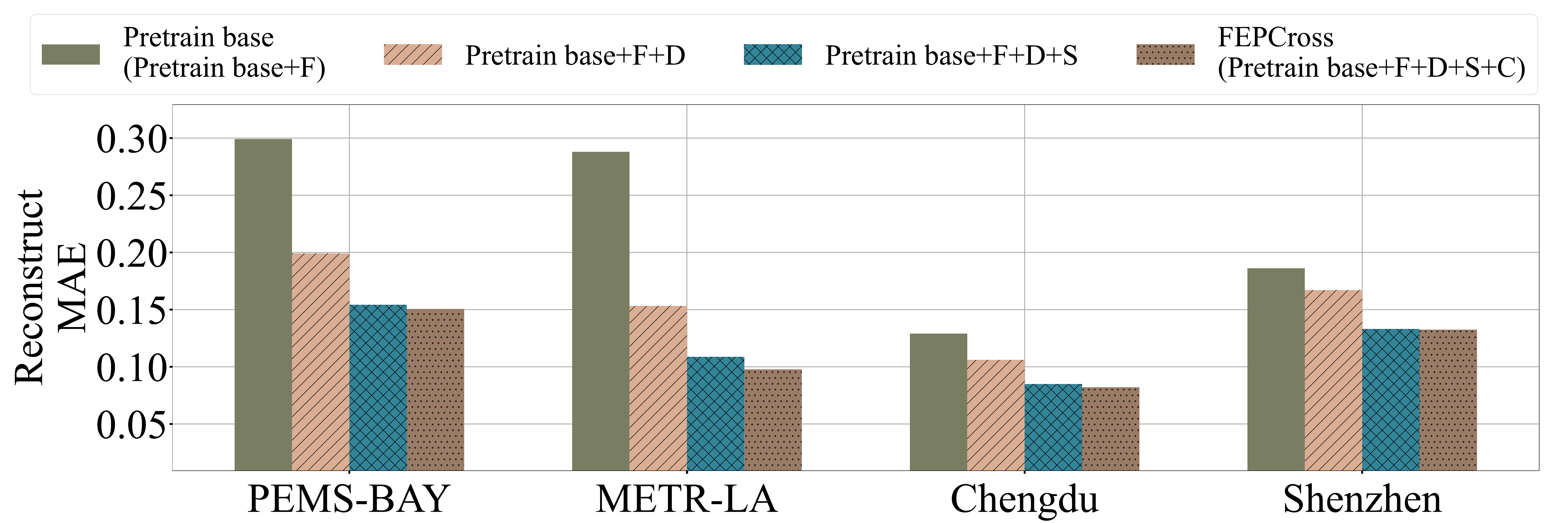}
    \caption{Reconstruction Analysis. The MAE of the reconstruction on the time domain is shown.
    Pretrain-base and Pretrain base+F have the same performance since they both individually reconstruct the data of the time domain. FEPCross, i.e., Pretrain base+F+D+S+C, is the final version.
    }
    \label{fig:pretrain_ablation}
\end{figure}

In this section, we aim to investigate the contribution of each module proposed in this paper.
We test the performance of six variations of FEPCross by sequentially adding modules in pre-training and fine-tuning on METR-LA and Chengdu datasets.
In the evaluation of the pre-training modules, all fine-tuning modules are added, and vice versa.
(1) \textbf{Pretrain base} only uses one encoder that takes the time domain as input.
(2) \textbf{Pretrain base+F(requency)} uses three individual encoders that take the time, amplitude, and phase domains as input.
(3) \textbf{Pretrain base+F+D(omain)} adds the Cross-domain Aggregator to fuse the information of three domains.
(4) \textbf{Pretrain base+F+D+S (patial)} adds the Cross-space Aggregator to fuse the spatial information.
(5) \textbf{Finetune base} directly uses the full pre-training modules and no fine-tune modules to generate predictions.
(6) \textbf{Finetune base+M} adds the momentum graph module.
(7) \textbf{FEPCross} is the final version with full modules of pre-training and fine-tuning. It can also be represented as \textbf{Pretrain base+F+D+S+C(ontrastive)} or \textbf{Finetune base+M+A(ugmented)}.
The result is shown in Table~\ref{tab:ablation}.
By incrementally incorporating the frequency-related modules from the pre-training, a discernible improvement in performance can be observed.
This observation demonstrates that the model effectively captures and learns city-invariant patterns in the frequency domain when addressing the cross-city problem, aligning with the intuition that the frequency domain is similar across different cities.
Furthermore, the inclusion of the momentum graph and training data enrichment leads to a notable enhancement in performance.
This indicates that the incorporation of these modules in the fine-tuning stage imparts robustness to the model and helps mitigate the risk of overfitting the few-shot data.

\section{RQ3: Reconstruction Analysis}

This section presents an investigation into the enhancement of the expressiveness of the framework through the proposed frequency domain modules. 
Unlike our previous study in RQ2, we shift our focus to evaluating the time-domain reconstruction error during the pre-training stage rather than the final few-shot forecasting error.
If the encoder exhibits more expressiveness, the time-domain reconstruction error should be lower.
Here, we consider four pre-training variations mentioned in RQ2 and evaluate their reconstruction performance in the time domain.
The result is shown in Fig.~\ref{fig:pretrain_ablation}.
We can see that sequentially adding the modules could lead to a better reconstruction error.
Moreover, adding the cross-domain (\textbf{D}) aggregator leads to a significant improvement, which demonstrates that the fusion of information of frequency domains contributes to the reconstruction of the time domain.

\section{RQ4: Cross Domain Analysis}

\begin{figure}[!t]
    \centering
    \includegraphics[width=0.80\linewidth]{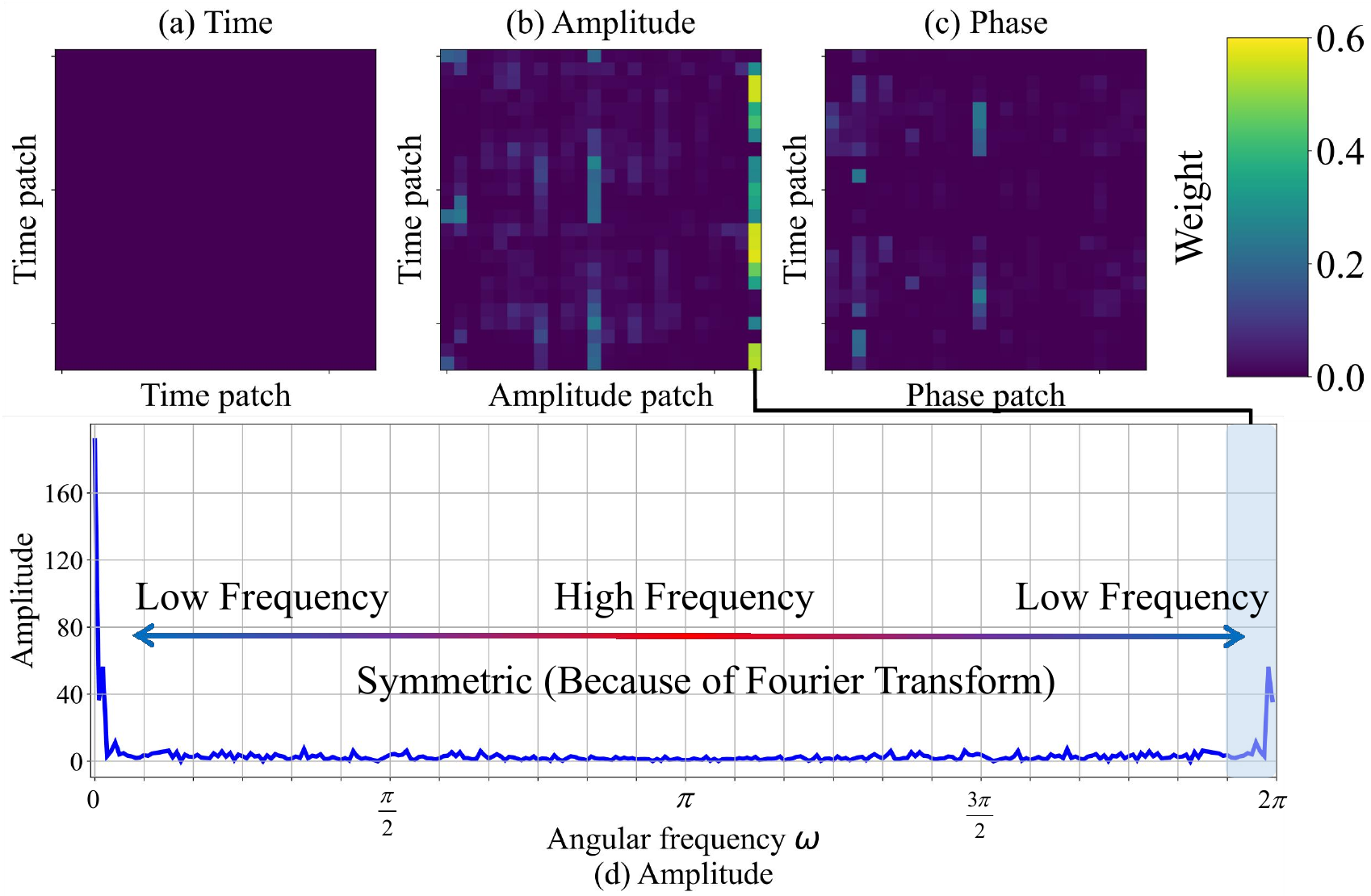}
    \caption{The attention map to reconstruct the time domain.}
    \label{fig:attention}
\end{figure}

In this section, we aim to investigate the effect that the amplitude and phase domain plays in the reconstruction of the time domain.
To visualize this effect, we present the attention map of the Cross-Domain Aggregator during the inference stage, as depicted in Fig.~\ref{fig:attention}.
Upon analyzing the attention map, we observe that the amplitude domain, particularly the low-frequency component, plays a significant role in the reconstruction of the time domain. 
In contrast, the time domain itself contributes less to the process.
These findings indicate that the Cross-Domain Aggregator effectively incorporates information from the frequency domain during the reconstruction pre-training. 
Consequently, the time domain receives a substantial amount of valuable information from the frequency domain.
This wealth of information enables the training of a robust and effective encoder, which in turn facilitates accurate traffic forecasting in the target city.

\section{Conclusion}
In this paper, we investigate the frequency-enhanced pre-training for the cross-city few-shot traffic forecasting problem.
We propose the FEPCross framework to incorporate the frequency domain into the forecasting in the target city.
Extensive experiments show that our proposed method not only outperforms other baselines but also exhibits good characteristics such as cross-domain auxiliary information sharing.
\section*{Acknowledgment}
This work was sponsored by National Key Research and Development Program of China under Grant No.2022YFB3904204, National Natural Science Foundation of China under Grant No. 62102246, 62272301, and Provincial Key Research and Development Program of Zhejiang under Grant No. 2021C01034. Part of the work was done when the students were doing internships at Yunqi Academy of Engineering.
\section*{Ethical Statement}
The data used in this paper (Chengdu, Shenzhen, PEMS-BAY, and METR-LA) are public open-sourced benchmark datasets, which are wildly used in academic research.
No personally identifiable information was obtained and people can not infer personal information through the data.
The potential use of this work is traffic knowledge transfer across cities, and this work could benefit the downstream applications of developing cities with little traffic data.
This work is not potentially a part of policing or military work.
The authors of this paper are committed to ethical principles and guidelines in conducting research and have taken measures to ensure the integrity and validity of the data.
The use of the data in this study is in accordance with ethical standards and is intended to advance knowledge in the field of cross-city few-shot traffic forecasting.

\clearpage
\bibliographystyle{splncs04}

\end{document}